\documentclass[conference]{IEEEtran}
\IEEEoverridecommandlockouts

\usepackage{amsmath,amssymb,amsfonts}
\usepackage{algorithmic}
\usepackage{acronym}
\usepackage{graphicx}
\usepackage{textcomp}
\usepackage{subcaption}
\usepackage{xcolor}
\def\BibTeX{{\rm B\kern-.05em{\sc i\kern-.025em b}\kern-.08em
    T\kern-.1667em\lower.7ex\hbox{E}\kern-.125emX}}

\usepackage[backend=biber, sorting=none, style=numeric-comp, maxnames=5]{biblatex}
\DeclareSourcemap{
  \maps[datatype=bibtex]{
    \map[overwrite=true]{
      \step[fieldset=url, null]
      \step[fieldset=eprint, null]
    }
  }
}

  \graphicspath{{img/} {figures/} {figures_SM/} {./}}
\DeclareGraphicsExtensions{.pdf,.png,.svg,.jpg,.mps}
    \acrodef{ADC}[ADC]{Analog-to-Digital Converter}
\acrodef{ADEXP}[AdExp-IF]{Adaptive Exponential Integrate-and-Fire}
\acrodef{ADM}[ADM]{Asynchronous Delta Modulator}
\acrodef{AE}[AE]{Address-Event}
\acrodef{AER}[AER]{Address-Event Representation}
\acrodef{AEX}[AEX]{AER EXtension board}
\acrodef{AFE}[AFE]{Analog Front-End}
\acrodef{AFM}[AFM]{Atomic Force Microscope}
\acrodef{AGC}[AGC]{Automatic Gain Control}
\acrodef{AI}[AI]{Artificial Intelligence}
\acrodef{AMDA}[AMDA]{AER Motherboard with D/A converters}
\acrodef{AMPA}[AMPA]{$\alpha$-Amino-3-hydroxy-5-methyl-4-isoxazolepropionic Acid}
\acrodef{ANN}[ANN]{Artificial Neural Network}
\acrodef{API}[API]{Application Programming Interface}
\acrodef{APMOM}[APMOM]{Alternate Polarity Metal On Metal}
\acrodef{ARM}[ARM]{Advanced RISC Machine}
\acrodef{ASIC}[ASIC]{Application Specific Integrated Circuit}
\acrodef{BCM}[BMC]{Bienenstock-Cooper-Munro}
\acrodef{BD}[BD]{Bundled Data}
\acrodef{BEOL}[BEOL]{Back-end of Line}
\acrodef{BG}[BG]{Bias Generator}
\acrodef{BMI}[BMI]{Brain-Machince Interface}
\acrodef{BTB}[BTB]{Band-to-Band tunnelling}
\acrodef{bpm}[bpm]{Beats per Minute}
\acrodef{CA}[CA]{Cortical Automaton}
\acrodef{CAD}[CAD]{Computer Aided Design}
\acrodef{CAM}[CAM]{Content Addressable Memory}
\acrodef{CAVIAR}[CAVIAR]{Convolution AER Vision Architecture for Real-Time}
\acrodef{CCN}[CCN]{Cooperative and Competitive Network}
\acrodef{CDR}[CDR]{Clock-Data Recovery}
\acrodef{CFC}[CFC]{Current to Frequency Converter}
\acrodef{CHP}[CHP]{Communicating Hardware Processes}
\acrodef{CMIM}[CMIM]{Metal-Insulator-Metal Capacitor}
\acrodef{CML}[CML]{Current Mode Logic}
\acrodef{CMOL}[CMOL]{Hybrid CMOS nanoelectronic circuits}
\acrodef{CMOS}[CMOS]{Complementary Metal-Oxide-Semiconductor}
\acrodef{CNN}[CNN]{Convolutional Neural Network}
\acrodef{CNS}[CNS]{central Nervous System}
\acrodef{COTS}[COTS]{Commercial Off-The-Shelf}
\acrodef{CPG}[CPG]{Central Pattern Generator}
\acrodef{CPLD}[CPLD]{Complex Programmable Logic Device}
\acrodef{CPU}[CPU]{Central Processing Unit}
\acrodef{CSM}[CSM]{Cortical State Machine}
\acrodef{CSP}[CSP]{Constraint Satisfaction Problem}
\acrodef{CTXCTL}[CTXCTL]{CortexControl}
\acrodef{CV}[CV]{Coefficient of Variation}
\acrodef{DAC}[DAC]{Digital to Analog Converter}
\acrodef{DAS}[DAS]{Dynamic Auditory Sensor}
\acrodef{DAVIS}[DAVIS]{Dynamic and Active Pixel Vision Sensor}
\acrodef{DBN}[DBN]{Deep Belief Network}
\acrodef{DBS}[DBS]{Deep Brain Stimulation}
\acrodef{DFA}[DFA]{Deterministic Finite Automaton}
\acrodef{DIBL}[DIBL]{Drain-Induced Barrier-Lowering}
\acrodef{DI}[DI]{Delay Insensitive}
\acrodef{divmod3}[DIVMOD3]{Divisibility of a number by three}
\acrodef{DMA}[DMA]{Direct Memory Access}
\acrodef{DNF}[DNF]{Dynamic Neural Field}
\acrodef{DNN}[DNN]{Deep Neural Network}
\acrodef{DOF}[DOF]{Degrees of Freedom}
\acrodef{DPE}[DPE]{Dynamic Parameter Estimation}
\acrodef{DPI}[DPI]{Differential Pair Integrator}
\acrodef{DRAM}[DRAM]{Dynamic Random Access Memory}
\acrodef{DR}[DR]{Dual Rail}
\acrodef{DRRZ}[DR-RZ]{Dual-Rail Return-to-Zero}
\acrodef{DSP}[DSP]{Digital Signal Processor}
\acrodef{DVS}[DVS]{Dynamic Vision Sensor}
\acrodef{DYNAP-SE}[DYNAP-SE]{Dynamic Neuromorphic Asynchronous Processor}
\acrodef{EBL}[EBL]{Electron Beam Lithography}
\acrodef{ECG}[ECG]{Electrocardiography}
\acrodef{ECoG}[ECoG]{Electrocorticography}
\acrodef{EDA}[EDA]{Electrodermal activity}
\acrodef{EDVAC}[EDVAC]{Electronic Discrete Variable Automatic Computer}
\acrodef{EEG}[EEG]{Electroencephalography}
\acrodef{EI}[EI]{Excitatory-Inhibitory}
\acrodef{EIN}[EIN]{Excitatory-Inhibitory Network}
\acrodef{EM}[EM]{Expectation Maximization}
\acrodef{EMG}[EMG]{Electromyography}
\acrodef{EOG}[EOG]{Electrooculogram}
\acrodef{EPSC}[EPSC]{Excitatory Post-Synaptic Current}
\acrodef{EPSP}[EPSP]{Excitatory Post-Synaptic Potential}
\acrodef{EZ}[EZ]{Epileptogenic Zone}
\acrodef{FDSOI}[FDSOI]{Fully-Depleted Silicon on Insulator}
\acrodef{FET}[FET]{Field-Effect Transistor}
\acrodef{FFT}[FFT]{Fast Fourier Transform}
\acrodef{FI}[F-I]{Frequency--Current}
\acrodef{FMA}[FMA]{Floating Microelectrode Array}
\acrodef{FNN}[FNN]{Feed-forward Neural Network}
\acrodef{FPGA}[FPGA]{Field Programmable Gate Array}
\acrodef{FR}[FR]{Fast Ripple}
\acrodef{FSA}[FSA]{Finite State Automaton}
\acrodef{FSM}[FSM]{Finite State Machine}
\acrodef{GABA}[GABA]{$\gamma$-Aminobutanoic Acid}
\acrodef{GIDL}[GIDL]{Gate-Induced Drain Leakage}
\acrodef{GLM}[GLM]{Generalized Linear Model}
\acrodef{GOPS}[GOPS]{Giga-Operations per Second}
\acrodef{GPIO}[GPIO]{General Purpose I/O}
\acrodef{GPU}[GPU]{Graphical Processing Unit}
\acrodef{GT}[GT]{Ground Truth}
\acrodef{GUI}[GUI]{Graphical User Interface}
\acrodef{HAL}[HAL]{Hardware Abstraction Layer}
\acrodef{HFO}[HFO]{High Frequency Oscillation}
\acrodef{HH}[H\&H]{Hodgkin \& Huxley}
\acrodef{HMM}[HMM]{Hidden Markov Model}
\acrodef{HR}[HR]{Heart Rate}
\acrodef{HRS}[HRS]{High-Resistive State}
\acrodef{HSD}[HSD]{Honest Significant Difference} 
\acrodef{HSE}[HSE]{Handshaking Expansion}
\acrodef{HW}[HW]{Hardware}
\acrodef{hWTA}[hWTA]{Hard Winner-Take-All}
\acrodef{HRV}[HRV]{hearth rate variability}
\acrodef{IC}[IC]{Integrated Circuit}
\acrodef{ICA}[ICA]{Indipendent Component Analysis}
\acrodef{ICT}[ICT]{Information and Communication Technology}
\acrodef{iEEG}[iEEG]{Intracranial Electroencephalography}
\acrodef{IF2DWTA}[IF2DWTA]{Integrate \& Fire 2-Dimensional WTA}
\acrodef{IF}[I\&F]{Integrate-and-Fire}
\acrodef{IFSLWTA}[IFSLWTA]{Integrate \& Fire Stop Learning WTA}
\acrodef{IMU}[IMU]{Inertial Measurement Unit}
\acrodef{INCF}[INCF]{International Neuroinformatics Coordinating Facility}
\acrodef{INI}[INI]{Institute of Neuroinformatics}
\acrodef{IO}[I/O]{Input/Output}
\acrodef{IoT}[IoT]{Internet of Things}
\acrodef{IP}[IP]{Intellectual Property}
\acrodef{IPSC}[IPSC]{Inhibitory Post-Synaptic Current}
\acrodef{IPSP}[IPSP]{Inhibitory Post-Synaptic Potential}
\acrodef{ISI}[ISI]{Inter-Spike Interval}
\acrodef{JFLAP}[JFLAP]{Java - Formal Languages and Automata Package}
\acrodef{LEDR}[LEDR]{Level-Encoded Dual-Rail}
\acrodef{LFP}[LFP]{Local Field Potential}
\acrodef{LIFE}[LIFE]{Longitudinal Intrafascicular Electrodes}
\acrodef{LIF}[LIF]{Leaky Integrate-and-Fire}
\acrodef{LLC}[LLC]{Low Leakage Cell}
\acrodef{LMS}[LMS]{Least Mean Squares}
\acrodef{LNA}[LNA]{Low-Noise Amplifier}
\acrodef{LPF}[LPF]{Low Pass Filter}
\acrodef{LR}[LR]{Logistic Regression}
\acrodef{LRS}[LRS]{Low-Resistive State}
\acrodef{LSM}[LSM]{Liquid State Machine}
\acrodef{LTD}[LTD]{Long Term Depression}
\acrodef{LTI}[LTI]{Linear Time-Invariant}
\acrodef{LTP}[LTP]{Long Term Potentiation}
\acrodef{LTU}[LTU]{Linear Threshold Unit}
\acrodef{LUT}[LUT]{Look-Up Table}
\acrodef{LVDS}[LVDS]{Low Voltage Differential Signaling}
\acrodef{MCMC}[MCMC]{Markov-Chain Monte Carlo}
\acrodef{MAE}[MAE]{Mean Absolute Error}
\acrodef{MEA}[MEA]{Multielectrode Arrays}
\acrodef{MEMS}[MEMS]{Micro Electro Mechanical System}
\acrodef{MFR}[MFR]{Mean Firing Rate}
\acrodef{MIM}[MIM]{Metal Insulator Metal}
\acrodef{ML}[ML]{Machine Learning}
\acrodef{MLP}[MLP]{Multilayer Perceptron}
\acrodef{monoNSM}[monoNSM]{Monotonic Neural State Machine}
\acrodef{MOSCAP}[MOSCAP]{Metal Oxide Semiconductor Capacitor}
\acrodef{MOSFET}[MOSFET]{Metal Oxide Semiconductor Field-Effect Transistor}
\acrodef{MOS}[MOS]{Metal Oxide Semiconductor}
\acrodef{MRI}[MRI]{Magnetic Resonance Imaging}
\acrodef{NCS}[NCS]{Neuromorphic Cognitive Systems}
\acrodef{NDFSM}[NDFSM]{Non-deterministic Finite State Machine}
\acrodef{ND}[ND]{Noise-Driven}
\acrodef{NEF}[NEF]{Neural Engineering Framework}
\acrodef{NHML}[NHML]{Neuromorphic Hardware Mark-up Language}
\acrodef{NIL}[NIL]{Nano-Imprint Lithography}
\acrodef{NI}[NI]{Neural Interface}
\acrodef{NMDA}[NMDA]{\textit{N}-Methyl-\textsc{d}-aspartate}
\acrodef{NME}[NE]{Neuromorphic Engineering}
\acrodef{NN}[NN]{Neural Network}
\acrodef{nnNSM}[nnNSM]{Nearest Neighbors Neural State Machine}
\acrodef{NOC}[NoC]{Network-on-Chip}
\acrodef{NRZ}[NRZ]{Non-Return-to-Zero}
\acrodef{NSM}[NSM]{Neural State Machine}
\acrodef{OR}[OR]{Operating Room}
\acrodef{OTA}[OTA]{Operational Transconductance Amplifier}
\acrodef{PCB}[PCB]{Printed Circuit Board}
\acrodef{PCHB}[PCHB]{Pre-Charge Half-Buffer}
\acrodef{PCM}[PCM]{Phase Change Memory}
\acrodef{PC}[PC]{Personal Computer}
\acrodef{PDK}[PDK]{Process Design Kit}
\acrodef{PE}[PE]{Phase Encoding}
\acrodef{PFA}[PFA]{Probabilistic Finite Automaton}
\acrodef{PFC}[PFC]{Prefrontal Cortex}
\acrodef{PFM}[PFM]{Pulse Frequency Modulation}
\acrodef{PGA}[PGA]{Programmable Gain Amplifier}
\acrodef{PNI}[PNI]{Peripheral Nerve Interface}
\acrodef{PNS}[PNS]{Peripheral Nervous System}
\acrodef{PPG}[PPG]{Photoplethysmography}
\acrodef{PR}[PR]{Production Rule}
\acrodef{PSC}[PSC]{Post-Synaptic Current}
\acrodef{PSD}[PSD]{Power spectral density}
\acrodef{PSP}[PSP]{Post-Synaptic Potential}
\acrodef{PSTH}[PSTH]{Peri-Stimulus Time Histogram}
\acrodef{PV}[PV]{Parvalbumin}
\acrodef{QDI}[QDI]{Quasi Delay Insensitive}
\acrodef{RAM}[RAM]{Random Access Memory}
\acrodef{RA}[RA]{Resected Area}
\acrodef{RDF}[RDF]{Random Dopant Fluctuation}
\acrodef{RELU}[ReLu]{Rectified Linear Unit}
\acrodef{RLS}[RLS]{Recursive Least-Squares}
\acrodef{RMSE}[RMSE]{Root Mean Square-Error}
\acrodef{RRMSE}[RRMSE]{Relative Root Mean Square Error}
\acrodef{RMS}[RMS]{Root Mean Square}
\acrodef{RNN}[RNN]{Recurrent Neural Network}
\acrodef{ROLLS}[ROLLS]{Reconfigurable On-Line Learning Spiking}
\acrodef{RRAM}[R-RAM]{Resistive Random Access Memory}
\acrodef{R}[R]{Ripple}
\acrodef{RBF}[RBF]{Radial basis function}
\acrodef{RISC}[RISC]{Reduced Instruction Set Computer}
\acrodef{RSA}[RSA]{Respiratory Sinus Arrhythmia}
\acrodef{SAC}[SAC]{Selective Attention Chip}
\acrodef{SAT}[SAT]{Boolean Satisfiability Problem}
\acrodef{SCI}[SCI]{Spinal Cord Injury}
\acrodef{SCX}[SCX]{Silicon CorteX}
\acrodef{SD}[SD]{Signal-Driven}
\acrodef{SEM}[SEM]{Spike-based Expectation Maximization}
\acrodef{SCR}[SCR]{Skin Conductance Response}
\acrodef{SLAM}[SLAM]{Simultaneous Localization and Mapping}
\acrodef{SNN}[SNN]{Spiking Neural Network}
\acrodef{SNR}[SNR]{Signal to Noise Ratio}
\acrodef{SOC}[SoC]{System-On-Chip}
\acrodef{SOI}[SOI]{Silicon on Insulator}
\acrodef{SOZ}[SOZ]{Seizure Onset Zone}
\acrodef{SP}[SP]{Separation Property}
\acrodef{SPI}[SPI]{Serial Peripheral Interface}
\acrodef{SRAM}[SRAM]{Static Random Access Memory}
\acrodef{SST}[SST]{Somatostatin}
\acrodef{STDP}[STDP]{Spike-Timing Dependent Plasticity}
\acrodef{STD}[STD]{Short-Term Depression}
\acrodef{STP}[STP]{Short-Term Plasticity}
\acrodef{STT-MRAM}[STT-MRAM]{Spin-Transfer Torque Magnetic Random Access Memory}
\acrodef{STT}[STT]{Spin-Transfer Torque}
\acrodef{SVM}[SVM]{Support Vector Machine}
\acrodef{SW}[SW]{Software}
\acrodef{sWTA}[sWTA]{soft Winner-Take-All}
\acrodef{TEMP}[TEMP]{Temperature}
\acrodef{TCAM}[TCAM]{Ternary Content-Addressable Memory}
\acrodef{TFT}[TFT]{Thin Film Transistor}
\acrodef{TIME}[TIME]{Transverse Intrafascicular Multichannel Electrode}
\acrodef{TLE}[TLE]{Temporal Lobe Epilepsy}
\acrodef{UEA}[UEA]{Utah Electrode Array}
\acrodef{USB}[USB]{Universal Serial Bus}
\acrodef{USEA}[USEA]{Utah Slanted Electrode Array}
\acrodef{VHDL}[VHDL]{VHSIC Hardware Description Language}
\acrodef{VHSIC}[VHSIC]{Very High Speed Integrated Circuits}
\acrodef{VIP}[VIP]{Vasoactive Intestinal Peptide}
\acrodef{VLSI}[VLSI]{Very Large Scale Integration}
\acrodef{VNS}[VNS]{Vagal Nerve Stimulation}
\acrodef{VOR}[VOR]{Vestibulo-Ocular Reflex}
\acrodef{VSA}[VSA]{Vector Symbolic Architecture}
\acrodef{WCST}[WCST]{Wisconsin Card Sorting Test}
\acrodef{WTA}[WTA]{Winner-Take-All}
\acrodef{XML}[XML]{eXtensible Mark-up Language}

\begin{document}

\title{Spiking Neural Networks for Mental Workload Classification with a Multimodal Approach}

\author{\IEEEauthorblockN{
Jiahui An\IEEEauthorrefmark{1}\IEEEauthorrefmark{2},
Sara Irina Fabrikant\IEEEauthorrefmark{3}
Giacomo Indiveri\IEEEauthorrefmark{2},
Elisa Donati\IEEEauthorrefmark{2}%
}
\vspace{0.2cm}

\IEEEauthorblockA{\IEEEauthorrefmark{1}Institute of Neuroinformatics, University of Zurich and ETH Zurich, Zurich, Switzerland\\ }
\IEEEauthorblockA{\IEEEauthorrefmark{2}Digital Society Initiative, University of Zurich, Zurich, Switzerland\\}

\IEEEauthorblockA{\IEEEauthorrefmark{3} Department of Geography and Digital Society Initiative, University of Zürich, Zürich, Switzerland\\ }

}

\maketitle

\begin{abstract}
  Accurately assessing mental workload is crucial in cognitive neuroscience, human-computer interaction, and real-time monitoring, as cognitive load fluctuations affect performance and decision-making.
  While \ac{EEG}-based machine learning (ML) models can be used to this end, their high computational cost hinders embedded real-time applications.
  Hardware implementations of spiking neural networks (SNNs) offer a promising alternative for low-power, fast, event-driven processing.

  This study compares hardware-compatible SNN models with various traditional ML ones, using an open-source multimodal dataset.
  Our results show that multimodal integration improves accuracy, with SNN performance comparable to the ML one, demonstrating their potential for real-time implementations of cognitive load detection.

  These findings position event-based processing as a promising solution for low-latency, energy-efficient workload monitoring, in adaptive closed-loop embedded devices that dynamically regulate cognitive demands.
\end{abstract}


\section{Introduction}
\label{sec:intro}
Mental workload (also called cognitive load) refers to the mental effort or resources a person uses to perform a task, distinct from the task's external demands~\cite{Puvsica_etal24}.

Mental workload classification plays a vital role in enhancing human-computer interaction, advancing cognitive neuroscience, and enabling real-time physiological monitoring.
Understanding and accurately assessing cognitive load is essential for optimizing system responsiveness, improving user experience, and ensuring safety in high-stakes environments~\cite{Dehais_etal20,Di_etal18,Kosch_etal23,Longo_etal18}.

Traditional approaches have primarily relied on \ac{ML}-based classifiers trained on brain signal datasets such as \acf{EEG} ones~\cite{Puvsica_etal24,Di_etal18,Hassan_etal24}.
However, these models often require extensive computational resources, limiting their deployment in real-time and low-power applications~\cite{O_etal13}.
Moreover, since these studies are focusing solely on \ac{EEG}, they overlook the potential benefits of multimodal physiological data integration~\cite{Hassan_etal24,Ding_etal20}.

In this study, we used an open-source multimodal dataset (EEG, heart rate variability, electrodermal activity, and skin temperature)~\cite{Anders_etal24}, which was evaluated only using a \ac{LR} model.
This limitation presented an opportunity for further exploration using alternative advanced classification techniques.
Therefore we developed an event-based approach that can be deployed onto neuromorphic chips to enable mental workload classification with low latency and low energy consumption.
Indeed, neuromorphic computing, inspired by the efficiency of biological neural networks, offers a promising alternative to conventional \ac{ML} models by leveraging \acp{SNN} for event-driven computation~\cite{Chicca_etal14, O_etal13}.

Unlike previous studies that solely relied on \ac{EEG} or other brain signals for mental workload classification, this work incorporates the analysis of \ac{EEG}, \ac{HRV}, \ac{EDA}, and \ac{TEMP} features.
The integration of multiple physiological signals can enhance classification performance by capturing a more comprehensive representation of cognitive load variations~\cite{Ding_etal20,Liu_etal23}.

While \ac{ML}-based classifiers such as \acp{SVM} and \ac{LR} have been widely used for mental workload classification, neuromorphic approaches based on \acp{SNN} remain largely unexplored.
This work addresses two key questions: (1) Can \acp{SNN} achieve performance comparable to state-of-the-art \ac{ML} models in mental workload classification? (2) Are brain signals alone sufficient to detect cognitive load changes, or is a multimodal approach essential for robust classification?

For the first time, we demonstrate the use of \ac{SNN} for cognitive load classification, building on previous successful results in biosignal processing.
Neuromorphic architectures have demonstrated the ability to detect anomalies in \ac{ECG} recordings~\cite{Bauer_etal19, Carpegna_etal24, DeLuca_etal24}, classify \ac{EMG} signals~\cite{Donati_etal19,Ma_etal20a, Ceolini_etal20}, and identify relevant biomarkers in \ac{EEG} data measured from epileptic patients~\cite{Sharifshazileh_etal21, Gallou_etal24}.
These findings suggest that \acp{SNN} offer a promising, biologically inspired alternative to traditional \ac{ML} approaches for cognitive load monitoring~\cite{Gong_etal23,Kumarasinghe_etal21,Burelo_etal22,Hadad_etal24,Yan_etal22}.

To address these questions, we systematically compared \ac{ML}-based methods with neuromorphic approaches, assessing their classification accuracy across different feature sets.
Our findings offer new insights into the potential of \acp{SNN} for mental workload monitoring, laying the foundation for real-time, energy-efficient implementations in wearable and embedded systems.

\section{Methods}
\label{sec:methods}
In this section, we describe our methodology for cognitive load classification using multimodal physiological signals.

\subsection{Data Description}
\label{ssec:data}
This study utilized an open-source dataset containing 315 hours of multimodal physiological recordings from 24 participants~\cite{Anders_etal24}.
Data were collected in both controlled laboratory settings and self-selected, uncontrolled environments.
However, we focused solely on the controlled sessions, as they involve standardized cognitive tasks that minimize external variability, ensuring consistency across participants and improving comparability and generalizability.

The dataset includes multiple modalities.
Brain signals were recorded using the Muse S \ac{EEG} headband at 256 Hz from five sensor locations: AF7 and AF8 at the frontal lobe, TP9 and TP10 at the temporal lobe, and FpZ as the reference channel following the EEG 10/20 international system (e.g., a standardized method for electrode placement based on scalp proportions).
Physiological signals were captured using the Empatica E4 wristband, which recorded \ac{PPG} at 64 Hz for \ac{HRV}, \ac{EDA} at 4 Hz to assess autonomic nervous system responses, and temperature at 4 Hz to monitor core body fluctuations.
Additionally, acceleration data were recorded at 32 Hz to account for motion-related artifacts.

Participants performed structured cognitive tasks designed to induce different levels of mental workload, including mental arithmetic, Stroop, N-back (two levels), and Sudoku.
Each task lasted 10 minutes, ensuring a balanced class distribution (Easy:Hard = 50:50).

For this study, we randomly selected 10 participants with complete controlled session data to balance computational efficiency and statistical robustness.
This subset provides sufficient data diversity while keeping the study focused and feasible.

\subsection{Preprocessing and Feature Extraction}
\label{ssec:preprocessing}
\ac{EEG} signals were bandpass-filtered between 0.5 and 50 Hz using a Butterworth filter to remove low-frequency drifts and high-frequency muscle artifacts, with an additional 50 Hz notch filter to eliminate power-line interference.
Baseline normalization was performed using an eye-closing session, and signals were average-referenced.
\ac{PSD} was computed using Welch's method for five frequency bands: $\delta$ (0.5–4 Hz), $\theta$ (4–7 Hz), $\alpha$ (8–12 Hz), $\beta$ (12–30 Hz), and $\gamma$ (30–50 Hz).
Asymmetry features were derived by subtracting the log-transformed spectral power of the right hemisphere from the left across all frequency bands.
Additionally, the $\theta/\alpha$ and $\alpha/\theta$ ratios were computed to enhance cognitive load estimation.

\ac{HRV} features were extracted from the \ac{PPG} signal using the NeuroKit2 package, including: time-domain metrics -- i.e.
mean RR interval (HRV-MeanNN), standard deviation of RR intervals (HRV-SDNN), and root mean square of successive differences (HRV-RMSSD) --, and frequency-domain measures -- normalized low-frequency power (HRV-LFn), normalized high-frequency power (HRV-HFn), and their ratio.
\ac{EDA} included the number of \ac{SCR} peaks and their mean amplitude, while temperature features were computed as the mean and standard deviation over time.

Variance thresholding resulted in most participants (9 out of 10) retaining 23 features, while one participant (UN\_105) retained 27 features due to higher physiological signal variability.
The four features removed for most participants were all related to \ac{EDA}—specifically, EDA\_Tonic\_SD, EDA\_Sympathetic, EDA\_SympatheticN, and EDA\_Autocorrelation.
Their removal suggests that these features exhibited low variance across participants and contributed minimally to cognitive load classification.

All signals were segmented using a 60-second sliding window with an 80\% overlap to effectively capture temporal dynamics.
Missing values were interpolated using the mean of neighboring values, and all features were standardized to achieve zero mean and unit variance.
Each 60-second segment was labeled as ``Easy'' (0) or ``Hard'' (1) based on the difficulty level of the corresponding cognitive task.

\begin{figure*}[htbp]
    \centering
    \includegraphics[width=0.95\textwidth]{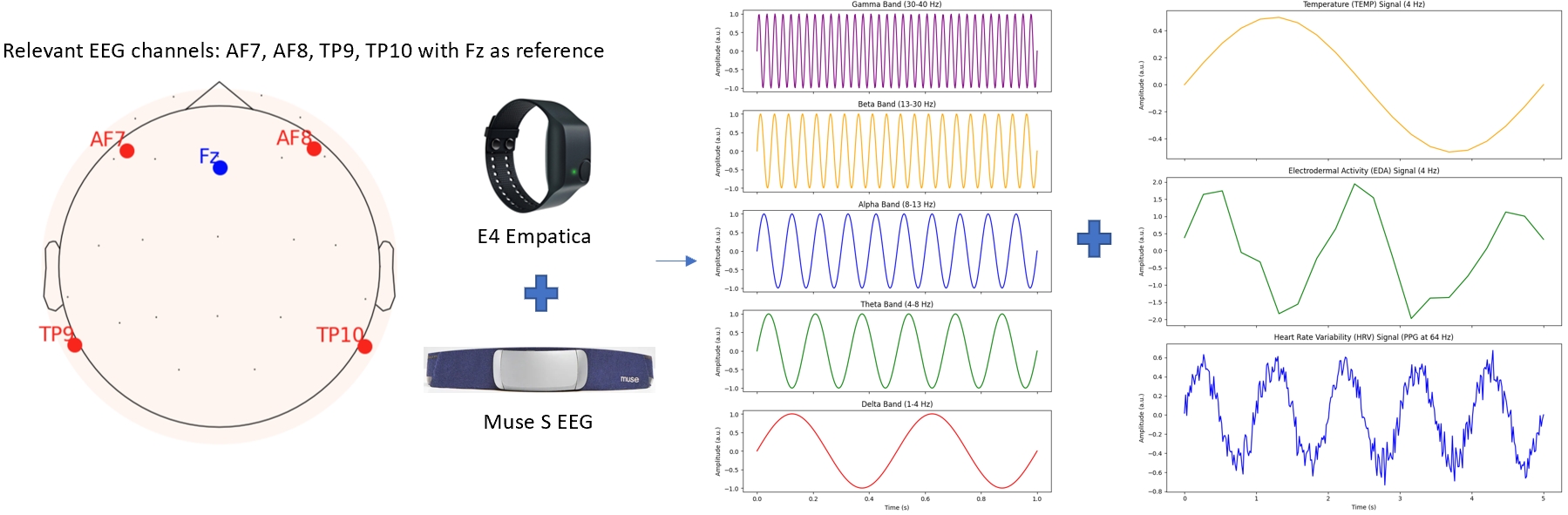}
    \caption{Schematic representation of the multimodal data acquisition system.
EEG signals were recorded using the Muse S headband at five sensor locations (AF7, AF8, TP9, TP10, and Fz as reference), while physiological signals, including heart rate variability, electrodermal activity, and temperature, were captured using the Empatica E4 wristband.
The recorded physiological signals were integrated for cognitive load classification after feature extraction.}
    \label{fig:system_overview}
\end{figure*}

Figure~\ref{fig:system_overview} provides an overview of the multimodal data acquisition setup, showing the EEG electrode placement and wearable sensors and extracted physiological signals used for cognitive load classification.

\subsection{Model Architecture and Training}
\label{ssec:model}
As a baseline, we implemented a \ac{LR} model with L2 regularization, optimized via grid search (over L1 and L2 penalties with solvers such as liblinear, saga and lbfgs) for each participant.
The data was partitioned into an 80/20 training/testing split, with stratified 5-fold cross-validation.
Compared to prior work using a similar dataset (which achieved 71\% accuracy) which were comparable to our \ac{LR} model (which achieved 78\%).
To further enhance performance, we explored more complex non-linear models, including \ac{MLP}, \ac{SVM} (Polynomial and RBF kernals), and \ac{SNN}, which are better suited for capturing complex relationships in multimodal physiological data.

\subsubsection{Models on EEG Features only}
To assess the effectiveness of \ac{EEG} features alone, we trained \ac{LR}, \ac{MLP}, and \ac{SVM} models using EEG-derived features.
\ac{EEG} preprocessing followed the pipeline described earlier, including standard scaling and variance threshold filtering to remove redundant features.
We implemented the \ac{LR} model as described in the baseline model above with \ac{EEG} features only.

For \ac{MLP}, we optimized the architecture and hyperparameters via grid search for each participant, which included one or two hidden layers (e.g., 50 neurons in a single layer, 100 neurons, two layers with 50 neurons each or two layers one with 50 neurons the other one with 100 neurons), \ac{RELU} or tanh activation, learning rates (0.001–0.01), and early stopping criteria.

We implemented \ac{SVM} models with linear, polynomial, and \ac{RBF} kernels were trained, with hyperparameter tuning including regularization strength \( C \), kernel-specific parameters (degree and gamma for polynomial and \ac{RBF} kernels), and class weights.
These \ac{EEG}-only models provided an initial benchmark before integrating multimodal physiological features.

To determine the optimal \ac{EEG} feature combination for cognitive load classification, we evaluated three feature sets: (1) \textit{Basic PSD}—including mean spectral power in five frequency bands: $\delta$ (1–4 Hz), $\theta$ (4–7 Hz), $\alpha$ (8–12 Hz), $\beta$ (12–30 Hz), $\gamma$ (30–50 Hz); (2) \textit{PSD + Asy}—which adds asymmetry measures (e.g., frontal-$\alpha$-asymmetry) to the Basic PSD; and (3) \textit{PSD + Asy + Ratios}—which further included power ratios (such as $\theta/\alpha$ and $\alpha/\theta$).

\subsubsection{Training ML-based Models with All Modalities}
\label{ssec:training}
With this analysis, we aim to investigate whether adding modalities improves classifier performance.
\ac{LR} models were optimized via grid search over regularization parameters, while \ac{MLP} tuning focused on hidden layer sizes, regularization parameter \(\alpha\), learning rate, and activation functions.
\ac{SVM} models retained the same tuning procedure as in \ac{EEG}-only experiments but now incorporated multimodal features.

\subsubsection{Training Spiking Neural Networks}
To further explore neuromorphic approaches for cognitive load classification, we implemented multiple \acp{SNN} using a spiking version of Python~\cite{Eshraghian_etal23}.

Our \acp{SNN} employed forward-pass, where input features were processed through the first layer (fc1) and its associated \ac{LIF} layer to produce spike outputs, which were then fed into a second \ac{LIF} layer (fc2).
Training follows two approaches:

\paragraph{Hybrid SNN (Bio-Inspired + Backpropagation)}
\begin{itemize}
    \item The fc1 weights are updated using the Adam optimizer via standard backpropagation.
    \item The fc2 weights are updated using a delta learning rule, where the error is computed as the difference between the target label and the sum of output spikes:
\end{itemize}

\begin{equation}
    \Delta W_{fc2} = \eta \cdot (S_{fc1}^\top \cdot \Delta),
    \label{eq:delta_w}
\end{equation}

where \( \eta \) is the learning rate, and \( S_{fc1}^\top \) represents the spike outputs from fc1.
The error (\(\Delta\)) is given by:

\begin{equation}
    \Delta = y - \sum S_{\text{out}}.
    \label{eq:error}
\end{equation}

Hyperparameter tuning for the network architecture was performed via grid search over the hidden layer size (set to 20 times the number of input features), learning rate (0.001, 0.01, or 0.1), number of epochs (10 or 20), and a batch size of 1, with early stopping applied after 5 epochs without improvement.
We employed stratified 5-fold cross-validation for each participant to select the best hyperparameter configuration based on mean validation accuracy and applied early stopping when performance did not improve over 5 consecutive epochs.
This approach—though it used backpropagation for fc1 updates—was intended for simulation with \texttt{snntorch} and served as a precursor to more biologically inspired models that would abandon backpropagation entirely for neuromorphic hardware implementation.

\paragraph{Bio-Inspired SNN}

We then implemented a second \ac{SNN} using one single layer training that allowed the implementation on a mixed analog/digital chip.
We first converted the continuous input features into spikes using \ac{LIF} neurons.
The spike encoding was achieved using parameters that mimic biological neurons: a resting potential of 0, a reset potential of \(-65.0\), a firing threshold of \(-50.0\), a membrane time constant \(\tau\), and a simulation time step of 1\,ms (\textit{dt}) across 40 steps.
The num\_steps which was defined by the dataset: the total recording for each participant lasted 40 minutes, and each feature was extracted based on a 1-minute' recording.
The \(\tau\) was optimized through grid search.

\begin{itemize}
    \item The fc1 weights remain fixed, initialized from a normal distribution with a mean of 0.1 and a standard deviation of 0.02.
    \item The fc2 weights are updated via the delta learning rule.
\end{itemize}

The \ac{LIF} neuron model followed the standard differential equation governing membrane potential dynamics:

\begin{equation}
    \tau \frac{dV(t)}{dt} = -(V(t) - V_{\text{rest}}) + X(t),
    \label{eq:lif}
\end{equation}

where \( V(t) \) is the membrane potential at time \( t \), \( V_{\text{rest}} \) is the resting membrane potential, and \( \tau \) is the membrane time constant.
Unlike conventional implementations, where \( I_{\text{in}}(t) \) represents an explicit input current, in this study, we define the input term \( X(t) \) as the input feature vector.
This means that instead of receiving a direct synaptic current, the LIF neuron directly integrates feature values as part of its membrane dynamics.

A spike is emitted when the membrane potential reaches the threshold \( V_{\text{th}} \), at which point the membrane potential is reset to \( V_{\text{reset}} \):

\begin{equation}
    V(t) = V_{\text{reset}}, \quad \text{if } V(t) \geq V_{\text{th}}.
    \label{eq:reset}
\end{equation}

\begin{figure}[htbp]
    \centering
    \includegraphics[width=0.95\columnwidth]{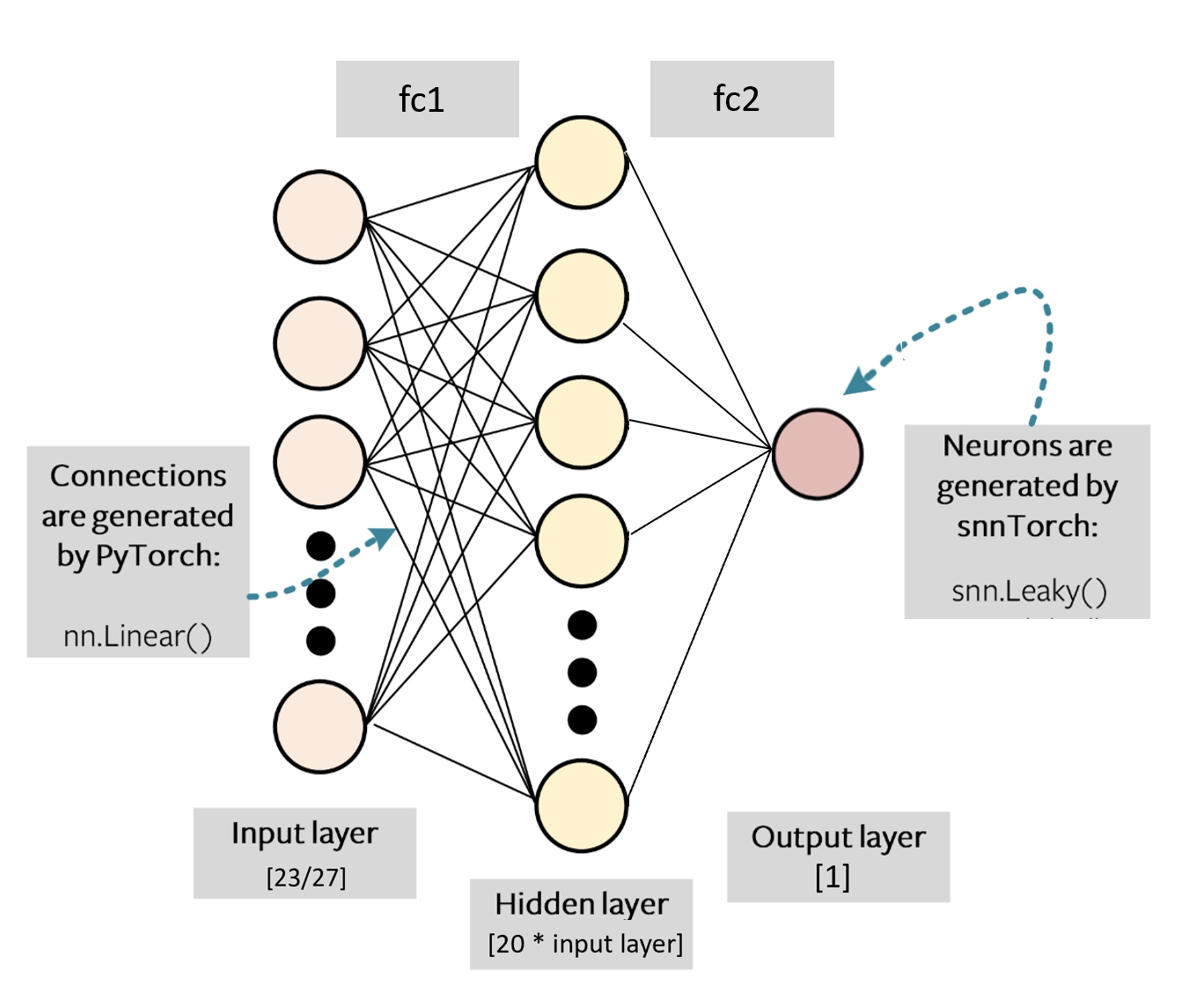}
    \caption{Schematic representation of the \ac{SNN} architecture.
The model consists of two fully connected layers (fc1 and fc2) with \ac{LIF} neurons.
The first layer (fc1) incorporates a connection probability parameter (p\_conn) and learnable beta in \ac{LIF} neurons, while the second layer (fc2) updates weights using a delta learning rule.}
    \label{fig:snn_architecture}
\end{figure}

The network architecture was similar to the previous \ac{SNN} but now applied \ac{LIF} neurons with a learnable beta parameter in both layers.
The second layer (fc2) remained fully connected, while the first layer (fc1) introduced a connection probability parameter (p\_conn).
The fc1 weights were initialized from a normal distribution with a mean of 0.1 and a standard deviation of 0.02 (i.e., 0.2 times the mean), following the same initialization for (p\_conn).
The optimal mean for fc1 weights and the best (p\_conn) distribution were determined via grid search.
A schematic representation is shown in Fig.~\ref{fig:snn_architecture}.

For classification, the output neuron's spiking activity is interpreted as follows: if the membrane potential reaches the threshold and a spike occurs, the network predicts Class 1; otherwise, it predicts Class 0.
The output neuron follows a threshold-based activation function with a default threshold of 1.0, ensuring discrete decision-making.

Additionally, we trained the bio-inspired \ac{SNN} on two different input configurations: EEG features only and all physiological modalities (EEG, HRV, EDA, TEMP).

\subsection{Evaluation Metrics}
\label{ssec:metrics}
Model performance was evaluated using four key metrics: accuracy, F1-score, precision, and recall.
These metrics were computed on an independent 20\% testing set.
To account for variability across participants, all performance metrics were averaged across stratified $k$-fold cross-validation splits ($k = 5$), providing robust estimates of classification performance.

To determine whether significant differences existed between model performances, we conducted a series of statistical tests on the classification accuracy across participants.
The selection of statistical tests was based on the distribution of model accuracies.

\subsubsection{Bootstrapping Procedure}
We applied a bootstrapping procedure by generating 1000 resampled accuracy distributions for each model based on their mean and standard deviation.
This resampling method enhanced statistical power and provided more reliable estimates of model performance variability.
The bootstrapped distributions were then used for subsequent statistical tests.

\subsubsection{Normality Test}
Before selecting appropriate statistical tests, we assessed whether the bootstrapped accuracy distributions for each model adhered to a normal distribution using the Shapiro-Wilk test.
If the normality assumption was satisfied ($p > 0.05$), we proceeded with parametric statistical tests, such as one-way ANOVA or Generalized Linear Model (GLM) for multiple model comparisons.
If the normality assumption was violated ($p \leq 0.05$), we opted for non-parametric alternatives, including Kruskal-Wallis test and Wilcoxon Signed-Rank test.

\subsubsection{Multiple Model Comparisons}
If normality was met, a one-way ANOVA was used to compare classification accuracies across different models.
If ANOVA produced a significant result ($p < 0.05$), post-hoc pairwise comparisons were conducted using Tukey's \ac{HSD} test.

If normality was not met, the Kruskal-Wallis test was used as a non-parametric test to compare classification accuracies across models.
If the Kruskal-Wallis test yielded a significant result ($p < 0.05$), pairwise comparisons were conducted using Conover's post-hoc test with Bonferroni correction to control for Type I errors.
To further interpret model differences, we computed Cliff's Delta for each pairwise comparison.
Cliff's Delta quantifies the effect size of the difference between two distributions, indicating whether a difference is practically meaningful in addition to being statistically significant and provides insights into the relative performance of different models.
A large effect size (\( |\delta| > 0.474 \)) indicates a strong practical difference, whereas a small effect size (\( 0.147 < |\delta| < 0.330 \)) suggests minor practical significance.
Effect sizes close to zero indicate negligible differences.

\section{Results}
\label{sec:results}

\subsection{Data Preprocessing and Feature Selection}
\label{ssec:feature_selection}
The dataset used in this study provided pre-extracted physiological features, as a result of preprocessing and feature selection, most participants (9 out of 10) retained 23 features, while one participant (UN\_105) retained 27 features.
The four features removed for most participants were all related to EDA, specifically EDA\_Tonic\_SD, EDA\_Sympathetic, EDA\_SympatheticN, and EDA\_Autocorrelation.



\subsection{EEG-Only ML Model Performance: Classification results for LR, MLP and SVM}
\label{ssec:only-EEG}
\begin{figure}[ht]
\centering
\begin{subfigure}[b]{0.95\columnwidth}
    \centering
    \includegraphics[width=\textwidth]{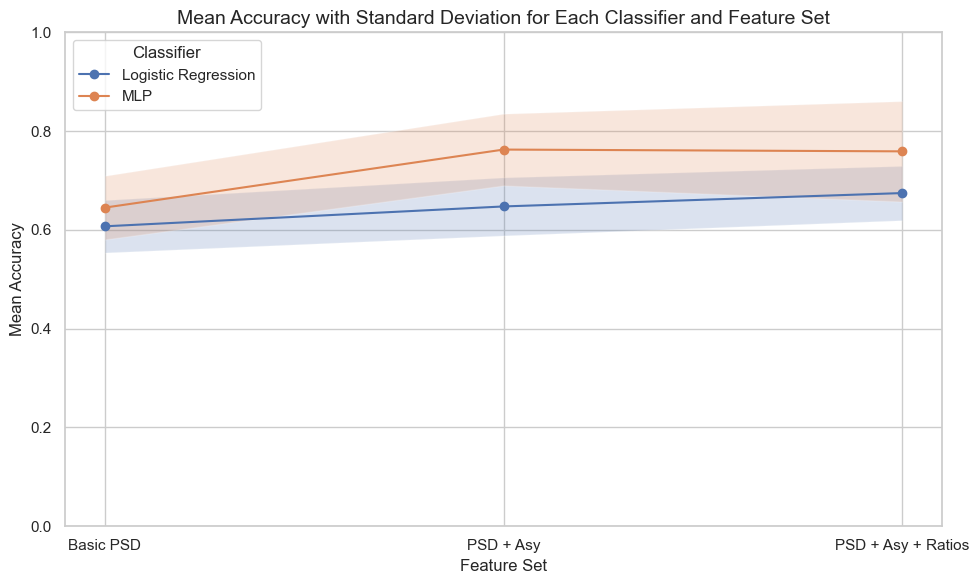}
    \label{fig:eeg_performance}
\end{subfigure}

\begin{subfigure}[b]{0.95\columnwidth}
    \centering
    \includegraphics[width=\textwidth]{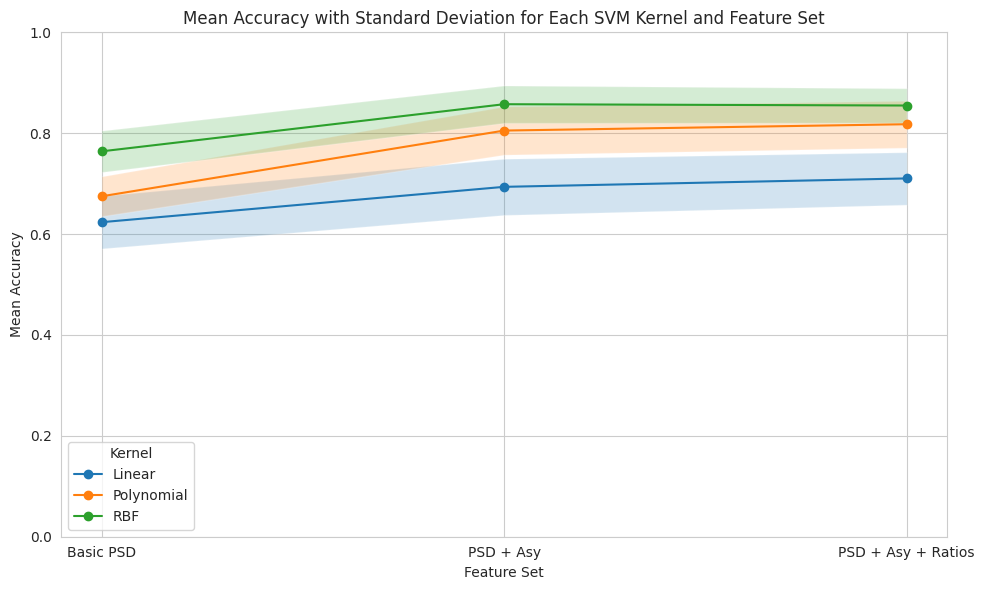}
    \label{fig:svm_performance}
\end{subfigure}
\caption{Comparison of EEG-only classification performance for different machine learning models.
Top plot: Logistic Regression and MLP across EEG feature sets.
Bottom plot: SVMs with different kernels.}
\label{fig:eeg_svm_combined}
\end{figure}

Figure~\ref{fig:eeg_svm_combined} illustrates the classification performance of \ac{LR}, \ac{MLP}, and \ac{SVM} across different \ac{EEG} feature sets.
\ac{MLP} consistently outperformed \ac{LR}, confirming that non-linear models benefit more from \ac{EEG}-based cognitive load classification.
With the Basic \ac{PSD} feature set, \ac{LR} achieved an accuracy of 60.7\% ± 5.3\%, while \ac{MLP} performed slightly better at 64.5\% ± 6.5\%.
Adding asymmetry features (\ac{PSD} + Asy) significantly improved both models, with \ac{LR} reaching 64.8\% ± 8.6\% and \ac{MLP} increasing to 76.3\% ± 7.3\%, marking the highest performance gain.
However, adding power ratio features (PSD + Asy + Ratios) slightly improved \ac{LR} (67.5\% ± 5.5\%) but caused a minor performance drop for \ac{MLP} (75.9\% ± 10.2\%), possibly due to feature redundancy or increased feature dimensionality.

Linear \ac{SVM} performed comparably to \ac{LR}, achieving 69.4\% accuracy on \ac{PSD} + Asy features.
However, non-linear kernels led to considerable improvements, with \ac{SVM} Polynomial achieving 81.8\% accuracy and \ac{RBF} \ac{SVM} reaching the highest performance of 85.6\% using the same feature set.
These results suggested that non-linear transformations in SVMs effectively capture cognitive load patterns in \ac{EEG} data.
Compared to \ac{MLP} (76.3\% accuracy on \ac{PSD} + Asy), the \ac{RBF} \ac{SVM} exhibited superior performance, indicating that well-tuned non-linear kernels might generalize better than neural networks in this dataset.

\subsection{Multimodal Model Performance}
\label{ssec:all_results}
The classification results, summarized in Table~\ref{tab:ml_performance}, demonstrated that incorporating multimodal data improved classification performance across all models.

Compared to the \ac{EEG}-only models, multimodal \ac{LR} accuracy increased from 67.5\% (\ac{EEG}) to 77.9\%, while \ac{MLP} performance improved from 76.3\% (\ac{EEG}) to 93.5\%.
This suggests that \ac{HRV}, \ac{EDA}, and \ac{TEMP} signals indeed contribute to cognitive load classification.

Among \ac{SVM} classifiers, \ac{SVM} \ac{RBF} (96.0\%) and \ac{SVM} Polynomial (94.0\%) outperformed and \ac{SVM} Linear (79.0\%).
The superior performance of \ac{RBF} and Polynomial kernels confirmed that non-linear models benefited from the additional physiological signals.

Compared to previous literature, where \ac{LR} achieved 71\% accuracy using all participants, our multimodal models showed significant improvement, particularly in \ac{MLP} and non-linear kernel methods \ac{SVM}.
This highlighted the effectiveness of multimodal feature fusion for cognitive load classification.

\begin{table}[ht]
\caption{Performance comparison of various machine learning models using multimodal features.}
\begin{center}
\begin{tabular}{|c|c|c|c|c|}
\hline
\textbf{Model} & \textbf{Accuracy} & \textbf{F1-Score} & \textbf{Precision} & \textbf{Recall} \\
\hline
Literature LR & 0.710 & 0.690 & 0.740 & 0.710 \\
Our LR & 0.778 & 0.777 & 0.785 & 0.771 \\
MLP & 0.934 & 0.935 & 0.935 & 0.936 \\
SVM Linear & 0.790 & 0.780 & 0.850 & 0.800 \\
SVM Polynomial & 0.940 & 0.940 & 0.970 & 0.940 \\
SVM RBF & 0.960 & 0.960 & 0.990 & 0.960 \\
\hline
\end{tabular}
\label{tab:ml_performance}
\end{center}
\end{table}

\subsection{Spiking Neural Network Performance}
\label{ssec:snn_results}
As a result of the grid search, the optimal hybrid \ac{SNN} model consisted of two fully connected layers without biases: the fc1 mapped the input features to a hidden representation (sized at 20 times the input dimension), and fc2 outputs a single neuron prediction.
Both layers were followed by ac{LIF} neurons with \(\beta=0.8\), this value was tuned also by grid search.
Moreover, \ac{SNN} models were trained using a batch size of 1, a learning rate of 0.001, and over 20 epochs.
We employed stratified 5-fold cross-validation for each participant to select the best hyperparameter configuration based on mean validation accuracy.

As for the biologically inspired \ac{SNN}, we used the same training process and network architecture except for the fc1 with a connection probability of 0.1.
Also, a learnable beta parameter was incorporated within the leaky activation function.
To transform continuous physiological features into spike representations, we applied \ac{LIF} spike encoding (\( V_{\text{rest}} = 0.0 \), \( V_{\text{reset}} = -65.0 \), \( V_{\text{th}} = -50.0 \), \( \tau = 29.0 \), \( dt = 1 \) ms, \( \text{num\_steps} = 40 \)).

Table~\ref{tab:snn_performance} presents the performance of the \ac{SNN} models trained for cognitive load classification.
The hybrid \ac{SNN}, which applied Adam optimization for fc1 and the Delta learning rule for fc2, achieved an accuracy of 88.4\% ± 3.5\%.
This suggested that hybrid \ac{SNN} architectures leveraging backpropagation could perform comparably to \ac{MLP} (93.4\% accuracy in Table~\ref{tab:ml_performance}).

\begin{figure}[ht]
\centerline{\includegraphics[width=0.95\columnwidth]{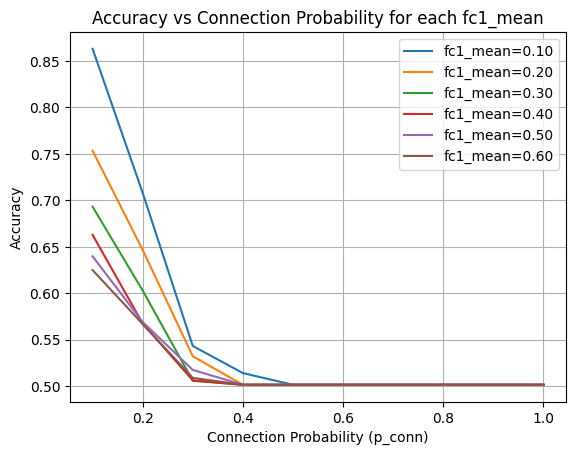}}
\caption{Accuracy vs.
Connection Probability (p\_conn) in Biologically Inspired SNN.}
\label{fig:p_conn_trend}
\end{figure}

The bio-inspired \ac{SNN}, which incorporated fc1 weight initialization from a normal distribution and a connection probability constraint (p\_conn), exhibited different performance trends based on input modality.
When trained only on \ac{EEG} features, accuracy dropped to 73.7\% ± 8.2\%.
However, when multimodal features were integrated, performance increased to 84.5\% ± 11.0\% surpassing the baseline \ac{LR} model.
These results indicated that biologically constrained \acp{SNN} could benefit significantly from multimodal integration.
Additionally, analysis of connection probability \(p\_{conn}\) trends (see Fig.~\ref{fig:p_conn_trend}) suggested that lower connectivity enhanced classification accuracy, reinforcing the importance of sparsity constraints in neuromorphic learning.

\begin{table}[ht]
\caption{Performance of Different SNN Architectures}
\begin{center}
\renewcommand{\arraystretch}{1.2}
\begin{tabular}{|l|c|c|c|c|}
\hline
\textbf{SNN Model} & \textbf{Accuracy} & \textbf{F1 Score} & \textbf{Precision} & \textbf{Recall} \\
\hline
Hybrid SNN & 0.885 & 0.881 & 0.902 & 0.866 \\
Bio-inspired SNN (EEG) & 0.735 & 0.722 & 0.771 & 0.735 \\
Bio-inspired SNN & 0.843 & 0.835 & 0.881 & 0.844 \\
\hline
\end{tabular}
\label{tab:snn_performance}
\end{center}
\end{table}

\subsection{Statistical Analysis of Model Performance}

\subsubsection{Bootstrapping and Normality Test Results}
A Shapiro-Wilk test was conducted to assess normality of bootstrapped accuracy distributions for each model, revealing that most models followed a normal distribution, including the hybrid \ac{SNN} ($W = 0.998, p = 0.204$), the bio-inspired \ac{SNN} trained with all modalities ($W = 0.999, p = 0.957$), and the bio-inspired \ac{SNN} trained on \ac{EEG} only ($W = 0.998, p = 0.491$).
However, the \ac{SVM} polynomial model ($W = 0.997, p = 0.041$) exhibited a statistically significant deviation from normality ($p < 0.05$).
Due to this, we opted for a non-parametric Kruskal-Wallis test to assess differences in model accuracy.

\subsubsection{Kruskal-Wallis Test Results}
The Kruskal-Wallis test yielded a highly significant result $(p < 0.0001)$, indicating that at least one model significantly outperformed or underperformed relative to the others.
To determine specific model differences, we conducted Conover's post-hoc analysis with Bonferroni correction.

\subsubsection{Post-hoc Analysis with Conover's Test and Cliff's Delta}
Table~\ref{tab:conover_results} presents the selected results of the pairwise Conover's test, including Bonferroni-corrected comparisons and Cliff's Delta effect sizes.
All reported comparisons are statistically significant.
Certain comparisons were omitted as they were not directly relevant to our research questions.

\begin{table}[ht]
\caption{Selected conover's Test Post-hoc Comparisons with Bonferroni corrected values and Cliff's Delta ($|\delta|$)}
\begin{center}
\renewcommand{\arraystretch}{1.2}
\begin{tabular}{|l|l|c|c|}
\hline
\textbf{Model 1} & \textbf{Model 2} & \textbf{Difference} & \textbf{$|\delta|$} \\
\hline
Literature LR & Our LR & -0.069 & 0.993 \\
Literature LR & Hybrid SNN & -0.174 & 1.000 \\
Literature LR & Bio-inspired SNN & -0.135 & 0.771 \\
Our LR & Hybrid SNN & -0.105 & 0.997 \\
Our LR & Bio-inspired SNN  & -0.066 & 0.461 \\
Hybrid SNN & Bio-inspired SNN & 0.039 & 0.268 \\
Bio-inspired SNN (EEG) & Bio-inspired SNN & -0.108 & 0.573 \\
SVM RBF & Hybrid SNN & 0.075 & 0.950 \\
SVM RBF & Bio-inspired SNN & 0.114 & 0.706 \\
\hline
\end{tabular}
\label{tab:conover_results}
\end{center}
\end{table}

\section{Discussion}
\label{sec:discussion}

\subsection{Impact of Feature Selection and EEG-Based Models}
\label{ssec:features_EEG}
The \ac{EEG}-only models demonstrated that feature selection significantly impacts classification accuracy.
The inclusion of asymmetry features (PSD + Asy) provided the highest accuracy gain, especially for non-linear models such as the \ac{MLP}.
However, adding power ratio features (PSD + Asy + Ratios) did not further enhance \ac{MLP} performance, indicating that feature redundancy may have introduced noise rather than improving classification.

\acp{SVM} with non-linear kernels (Polynomial and RBF) outperformed \ac{LR} and \ac{MLP} on \ac{EEG}-only features, achieving 85.6\% accuracy with the \ac{RBF} kernel.
This suggested that \ac{EEG}-based cognitive load classification benefited from non-linear transformations, which better captured complex neural activity patterns.

\ac{EEG}-only classification results highlighted the impact of feature selection and non-linear modeling.
Asymmetry features (\ac{PSD} + Asy) significantly improved performance, while power ratio features had mixed effects, enhancing \ac{LR} but slightly reducing \ac{MLP} accuracy.


However, \ac{EEG} alone remained insufficient for optimal classification, reinforcing the need for multimodal integration.

\subsection{Effectiveness of Multimodal Integration}
\label{ssec:multimodal_dis}
Our results confirmed that multimodal feature fusion significantly enhanced classification performance.
Compared to \ac{EEG}-only models, incorporating \ac{HRV}, \ac{EDA}, and \ac{TEMP} features improved model performance across all machine learning methods.


These findings indicated that physiological data from multiple modalities provided complementary information, reinforcing the need for multimodal integration in cognitive load assessment.

\subsection{Comparing Deep Learning and Neuromorphic Models}
\ac{SNN} models yielded interesting insights into neuromorphic approaches for cognitive load classification.
The hybrid \ac{SNN}, which leveraged backpropagation for fc1 and a Delta learning rule for fc2, achieved an accuracy of 88.4\%, significantly higher than baseline \ac{LR} model (77.8\%) and \ac{SVM} Linear (79.0\%) with large effect size.
This suggested that hybrid \ac{SNN} architectures could perform better with simple \ac{ML} models while maintaining biologically plausible learning dynamics.
Moreover, the hybrid \ac{SNN} showed higher accuracy than multimodal bio-inspired \ac{SNN} though with a relatively small effect size ($\Delta = 0.268$), indicating that their real-world performance remains comparable.

The bio-inspired \ac{SNN} exhibited different performance trends based on input modality.
When trained on \ac{EEG}-only features, its accuracy dropped to 73.5\%, reflecting the limited discriminative power of \ac{EEG} under biologically constrained learning rules.
However, when multimodal features were included, the bio-inspired \ac{SNN} reached 84.3\%.
This reinforced the idea that biologically inspired architectures could benefit significantly from multimodal integration.
Our SNN-based models exhibited higher accuracy variance (see Table~\ref{tab:mean_accuracy_variance}) compared to traditional \ac{ML} models, which demonstrated relatively lower accuracy fluctuations.
This discrepancy arises from the fact that \ac{ML} approaches underwent individualized hyperparameter tuning for each participant, whereas the \ac{SNN}-based models were trained using a fixed set of hyperparameters without participant-specific optimization.

\begin{table}[ht]
\caption{Mean Accuracy and Standard Deviation of Different Models.}
\begin{center}
\renewcommand{\arraystretch}{1.2}
\begin{tabular}{|c|c|c|}
\hline
\textbf{Model} & \textbf{Mean Accuracy} & \textbf{Standard Deviation} \\
\hline
Literature LR & 0.710 & 0.020 \\
Our LR & 0.778 & 0.015 \\
MLP & 0.934 & 0.017 \\
SVM Linear & 0.790 & 0.018 \\
SVM Polynomial & 0.940 & 0.016 \\
SVM RBF & 0.960 & 0.015 \\
Hybrid SNN & 0.885 & 0.034 \\
Bio-inspired SNN (EEG) & 0.735 & 0.079 \\
Bio-inspired SNN & 0.843 & 0.109 \\
\hline
\end{tabular}
\label{tab:mean_accuracy_variance}
\end{center}
\end{table}

\subsection{Statistical Validation of Model Performance}

Statistical analysis confirmed that the observed performance differences were significant.
The Kruskal-Wallis test revealed a significant effect of model type on classification accuracy ($p < 0.0001$).
However, practical post-hoc analysis showed that the significance varied depending on effect size and mean difference.

The multimodal bio-inspired \ac{SNN} significantly outperformed its \ac{EEG}-only variant, as indicated by a large effect size (\( |\delta| = 0.573\)), confirming that incorporating multiple physiological signals significantly enhanced model performance and reinforced the benefits of multimodal fusion.
Similarly, this model also showed medium-to-large effect sizes when compared to traditional machine learning models such as \ac{LR} and \ac{SVM} Linear, suggesting that neuromorphic models leveraging multimodal data could achieve superior accuracy compared to simpler \ac{ML} approaches.

\ac{SVM} \ac{RBF} outperformed both the hybrid \ac{SNN} (mean difference = 0.075, \( |\delta| = 0.950 \)) and bio-inspired \ac{SNN} (mean difference = 0.114, \( |\delta| = 0.706 \)).
These results suggested that well-tuned \ac{SVM} \ac{RBF} achieved better accuracy than neuromorphic approaches.

Furthermore, statistical analysis also confirmed that multimodal feature integration was essential for optimal performance in biologically inspired \acp{SNN}.
When using only \ac{EEG} data, performance significantly dropped.
However, when additional physiological signals were incorporated, the bio-inspired \ac{SNN} achieved much higher accuracy and was comparable to some well-tuned traditional \ac{ML} models.

Overall, these findings highlighted that while neuromorphic models showed promising improvements over certain \ac{ML} approaches, deep learning and well-optimized \ac{SVM} models remained strong competitors in cognitive load classification tasks.
These findings highlighted the potential of neuromorphic computing as an efficient alternative to traditional ML methods, particularly in real-time, low-latency, and low-power applications.

\section{Conclusion}
\label{sec:conclusion}
Our findings highlight the importance of multimodal feature integration, non-linear classification, and biologically constrained neuromorphic computing for cognitive load classification.
Future research should focus on optimizing \ac{SNN} architectures for real-time, low-power applications by leveraging raw signals to identify novel biomarkers for cognitive load detection and spike encoding.
Additionally, exploring biologically inspired learning rules could enhance \ac{SNN} performance, while neuromorphic hardware implementations should be investigated for real-time cognitive load monitoring.

\section{Acknowledgements}We acknowledge the financial support of the Digital Society Initiative (DSI) at University of Zurich.

\printbibliography

\end{document}